\documentclass[runningheads]{llncs}
\usepackage[T1]{fontenc}
\usepackage{graphicx}
\usepackage{amssymb}
\begin{document}
\title{CLIP-based Point Cloud Classification via Point Cloud to Image Translation}
%
%
\author{Shuvozit Ghose \inst{1} \and
Manyi Li\inst{2} \and
Yiming Qian\inst{1} and Yang Wang\inst{3}}
\authorrunning{Ghose et al.}
%
\institute{University of Manitoba, Canada \and
Shandong University, China \and
Concordia University, Canada}
\maketitle              
\begin{abstract}
Point cloud understanding is an inherently challenging problem because of the sparse and unordered structure of the point cloud in the 3D space. Recently, Contrastive Vision-Language Pre-training (CLIP) based point cloud classification model i.e. PointCLIP has added a new direction in the point cloud classification research domain. In this method, at first multi-view depth maps are extracted from the point cloud and passed through the CLIP visual encoder. To transfer the 3D knowledge to the network, a small network called an adapter is fine-tuned on top of the CLIP visual encoder. PointCLIP has two limitations. Firstly, the point cloud depth maps lack image information which is essential for tasks like classification and recognition. Secondly, the adapter only relies on the global representation of the multi-view features. Motivated by this observation, we propose a Pretrained Point Cloud to Image Translation Network (PPCITNet) that produces generalized colored images along with additional salient visual cues to the point cloud depth maps so that it can achieve promising performance on point cloud classification and understanding. In addition, we propose a novel viewpoint adapter that combines the view feature processed by each viewpoint as well as the global intertwined knowledge that exists across the multi-view features. The experimental results demonstrate the superior performance of the proposed model over existing state-of-the-art CLIP-based models on ModelNet10, ModelNet40, and  ScanobjectNN datasets.  

\end{abstract}
%
%
%
\section{Introduction}
\label{sec:intro}
Point cloud understanding refers to the process of extracting meaningful information from 3D point clouds, which are sets of 3D coordinates representing the surface geometry of objects or scenes. The goal of point cloud understanding is to analyze and interpret the data contained in the point cloud in order to understand the objects or scenes that it represents. Point cloud understanding has various applications in the real world, such as stereo reconstruction, indoor navigation, autonomous driving, augmented reality, and robotics perception etc. Although both 2D image understanding and point cloud understanding involve analyzing visual data, compared to the 2D image understanding \cite{radford2021learning}, 3D point cloud understanding \cite{meng2021towards} is more challenging. A 2D image consists of a dense and regular pixel array. In contrast, a 3D point cloud only consists of sparse and unordered points in the 3D space. Moreover, point clouds often lack the rich texture and image information available in 2D images.

\begin{figure}[t]
\begin{center}
\includegraphics[width=0.75\linewidth]{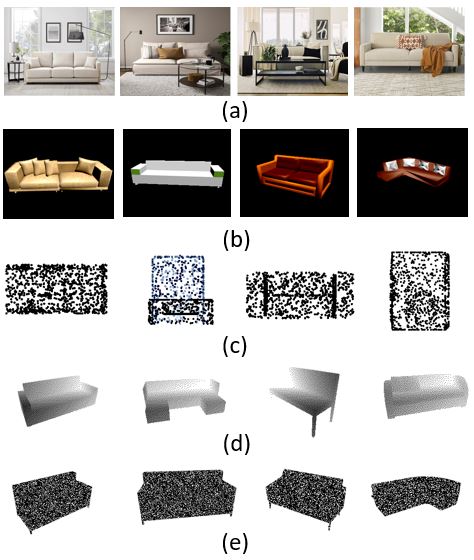} 
\end{center}

   \caption{Example of different image representations: (a) natural RGB images; (b) rendered RGB images; (c) point cloud depth maps; (d) 3D depth maps; (e) processed binary mask images.}

\label{fig:fig1}
\end{figure}

The success of deep learning in computer vision has also accelerated deep learning-based point cloud understanding and 3D-related research. While early deep learning methods had tried to propose some advanced architectures like PointNet \cite{qi2017pointnet}, PointNet++ \cite{qi2017pointnet++}, RSCNN \cite{liu2019relation}, DGCNN \cite{wang2019dynamic}, CurveNet \cite{xiang2021walk}, the success of Contrastive Language-Image Pre-Training (CLIP) model has added new direction in the context of computer vision. CLIP has several advantages over traditional deep learning methods. Firstly, the CLIP model is trained in a more generalizable manner, learning to associate images with natural language text in a way that can be applied to a wide range of downstream tasks. Whereas, traditional deep learning models are typically trained on specific tasks, such as image classification or object detection, and their performance can degrade significantly when applied to new, unseen tasks. Secondly, the CLIP model is trained on a large, unlabeled dataset of image-caption pairs, which does not require labeling efforts. On the other hand, traditional deep learning models often require large amounts of labeled training data to achieve good performance on a specific task. Finally, most importantly the CLIP model can be fine-tuned on new datasets and tasks with minimal additional training, making it a more flexible and adaptable solution to the downstream tasks compared to traditional deep learning models. Recently following the CLIP's success  on image   and natural language domain, several works have been proposed to generalize pre-trained clip to 3D recognition. Some of these works focus on designing a small adapter network to CLIP \cite{zhang2022pointclip} \cite{huang2022clip2point} and fine-tuning it for the downstream task. Other works focus on LLM-assisted 3D prompting and realistic shape projection \cite{zhu2022pointclip} and cross-modal training framework \cite{yan2022let} to bridge the gap between 2D image and point cloud.  In general, the pipeline is as follows. Given a point cloud, the point cloud is first projected as a depth map. The depth map is then processed by the pre-trained CLIP visual encoder \cite{radford2021learning}. A small network called an adapter is added and fine-tuned for the downstream task.  

Although these methods show some promising performance, they have certain limitations. It is due to the fact that the CLIP \cite{radford2021learning} is trained on RGB images whereas these models utilize point cloud depth maps for the point cloud understanding. Inherently, RGB images and depth maps are quite different from one another as depicted in Fig.~\ref{fig:fig1}. Point cloud depth maps represent depth information as a set of 3D points in space, with each point having an x, y, and z coordinate. This information is useful for 3D reconstruction and robotics navigation and manipulation. On the other hand, RGB images consist of red, green, and blue color channels, and each pixel in the image is represented by a combination of intensity values for these channels and captures color and texture information that is important for tasks like classification, recognition, and localization. In summary, the image information missing in the depth maps leads to the degrading performance  of the state-of-the-art CLIP-based \cite{radford2021learning} point cloud models. 

To transfer the image information to the CLIP-based point cloud models, one naive solution can be designing a network that maps depth maps to the corresponding natural RGB images. But, there does not exist any dataset that has depth maps and natural RGB image correspondence. However, there exists a dataset that has depth maps and rendered RGB image correspondence. In this direction, the next solution can be designing a network that maps depth maps to the corresponding rendered RGB images. Here the problems are three-fold. Firstly, CLIP is trained on natural RGB images. Rendered images differ from natural images in terms of realism, lighting, and Complexity depicted by Fig.~\ref{fig:fig1}~(a,b). Secondly, for a single depth map, there can be many possible corresponding rendered RGB images. For example, for a depth map of a sofa, the synthetic color changes in various parts as depicted in Fig.~\ref{fig:fig1}~(b) in multiple rendered image instances. Finally, a 3D model depth map differs from a point cloud depth map as showed by \cite{rothganger20063d}. A 3D model depth map typically represents depth information as a grayscale image shown in Fig.~\ref{fig:fig1}(d), with darker regions indicating greater distance from the viewer. In contrast, a point cloud depth map represents depth information as a set of 3D points in space, with each point having an $(x, y, z)$ coordinate as depicted by Fig.~\ref{fig:fig1}(c).  

In order to transfer image information to the CLIP \cite{radford2021learning} based point cloud model, we propose a novel Pretrained Point Cloud to Image Translation Network (PPCITNet) that produces generalized colored images along with additional salient visual cues to the point cloud depth maps. Here, the salient visual cues refer to additional color concentration to prominent or distinctive parts like an additional color concentration in the head and legs of a person (see Fig.~\ref{fig:visual}). The target of our PPCITNet is to provide image information to the CLIP \cite{radford2021learning} model so that it can achieve promising performance on point cloud classification and understanding. To pre-train this Point Cloud to Image Translation Network (PCITNet), we utilize the binary mask images of the rendered RGB images. Binary mask images and point cloud depth maps are similar geometrically because of discrete and compact representation. But visually, they are slightly different (see Fig.~\ref{fig:fig1}~(d, e)). To further bridge the gap, we preprocess the binary mask images by multiplying the binary image with a noise image to make the binary image sparse. The noise image is composed of $50\%$ white pixel and $50\%$ of black pixel sampled randomly. Through PPCITNet, the depth features of the point cloud can then be well aligned with the visual CLIP features. 

To further adapt our network to the few-shot learning, we proposed a novel viewpoint adapter that combines the local feature processed by each viewpoint as well as the global intertwined knowledge that existed across the multi-view features. In our opinion, the local viewpoint information is crucial for point cloud classification. For example, to classify the point cloud of 'airplane' the viewpoint that contains the wing information is more crucial than any other parts. In summary, the contributions of our paper are as follows. 1) We propose a novel Pretrained Point Cloud to Image Translation Network (PPCITNet) that transfers image information to the point cloud depth maps so that it can achieve promising performance on point cloud classification and understanding. 2) We propose a novel viewpoint adapter that combines the view feature processed by each viewpoint as well as the global intertwined knowledge existing across the multi-view features. 3) Our methods achieve state-of-the-art results on few-shot point cloud classification tasks on ModelNet10, ModelNet40, and ScanobjectNN.

\section{Related Works}\label{related}
\textbf{Deep Learning in Point Clouds.} Deep learning has revolutionized the field of point cloud classification and understanding. Categorically, deep learning-based models are divided into three sections, including multi-view based methods \cite{feng2018gvcnn}, volumetric-based methods \cite{li2016fpnn} and point-based methods \cite{qi2017pointnet}. Early works on deep learning primarily focused on multi-view-based methods~\cite{feng2018gvcnn}, where the 2D image models are utilized for point cloud classification. In volumetric-based methods \cite{li2016fpnn}, point clouds are treated as voxel data. 3D convolution-based models are used for classification and segmentation. The state-of-the-art models are point cloud-based methods~\cite{qi2017pointnet}, where the raw points are processed and passed through the model without any transformation. PointNet \cite{qi2017pointnet} is the first point-based model that has encoded each point with a multi-layer perception. PointNet++ \cite{qi2017pointnet++} further utilizes the max pooling operation to ensure permutation invariance. Recently, the success of CLIP for the downstream tasks on 2D has motivated the use of pre-trained CLIP for point cloud classification. Zhang et. al. \cite{zhang2022pointclip} propose PointCLIP which generalizes pre-trained CLIP to 3D recognition. 

\noindent\textbf{CLIP-based Point Cloud Models.} Recently several works have been proposed to generalize pre-trained Contrastive Language-Image Pre-Training (CLIP) to point cloud understanding tasks. For example, Zhang et. al. first proposed PointCLIP \cite{zhang2022pointclip} by extending the CLIP \cite{radford2021learning} for handling 3D point cloud data. In addition, they presented an inter-view adapter to capture the feature interaction between multiple views. In this direction, Zhu et. al.\cite{zhu2022pointclip} further introduced an efficient cross-modal adaptation method called PointCLIP V2 by proposing LLM-assisted 3D prompting and realistic shape projection. Next, Huang et. al. \cite{huang2022clip2point} presented a novel Dual-Path adapter and contrastive learning framework to transfer CLIP knowledge to the 3D domain. Yan et. al. \cite{yan2022let} presented PointCMT, an point cloud cross-modal training framework that utilized the merits of color-aware 2D images and textures to acquire more discriminative point cloud representation and formulated point cloud analysis as a knowledge distillation problem.

\section{Methodology} \label{proposed_method}
In this section, we first briefly revisit PointCLIP \cite{zhang2022pointclip} for few-shot 3D classification (Sec. \ref{revisit}). Then we introduce our Pretrained Point Cloud to Image Translation Network (PPCITNet) framework (Sec.\ref{pre}) that aligns image information to the point cloud depth map. Finally, we describe our proposed few-shot learning framework for few-shot point cloud classification (Sec. \ref{few}). The overall overview of our method is depicted in Fig.~\ref{fig:arch}.

\subsection{Revisit of PointCLIP} \label{revisit}
Similar to CLIP \cite{radford2021learning} which matches images and text by contrastive learning, PointCLIP \cite{zhu2022pointclip} consists of one visual encoder and a textual encoder. For $K$ class classification, PointCLIP uses a pre-defined template: "point cloud depth map of a big [CLASS]" and the textual encoder outputs $P \in R^{K \times C}$, where $C$ is the channel of the text embedding. To feed point clouds to the CLIP's visual encoder \cite{radford2021learning}, point clouds are first projected onto depth maps $\{f_1,f_2, \dots, f_M\}$. Here $M$ denotes the number of views and $f_i \in R^{H \times W \times C}$ denotes each view of the point cloud, where $H$ and $ W$ denote height and width respectively. Given the input $\{f_1,f_2, \dots, f_M\}$, the visual encoder in PointCLIP generates visual feature $\{F_1^I,F_2^I, \dots, F_M^I\}$, where $F_i^I \in R^{1 \times C}$ and $C$ is the channel dimension of the embedding.

\textbf{Zero-shot Classification:} In a zero-shot setup, there is no training stage. Each viewpoint generates a prediction by calculating the cosine similarity between the visual feature $F_i^I$ and the  textual feature $P^T$. The final prediction is the weighted sum of all viewpoint-wise predictions. Thus,

\begin{equation}
 O_i = F_i^I P^T,  i= 1,2 \dots N
\end{equation}

 \begin{equation}
 \widehat{y} =  softmax(\sum_{i=1}^{N} \alpha_i O_i \nonumber)
\end{equation}

where $\alpha_i$ is a hyper-parameter that describes the weighting importance of the view $i$.

\begin{figure*}[t]
\begin{center}
  \includegraphics[width=0.80\linewidth]{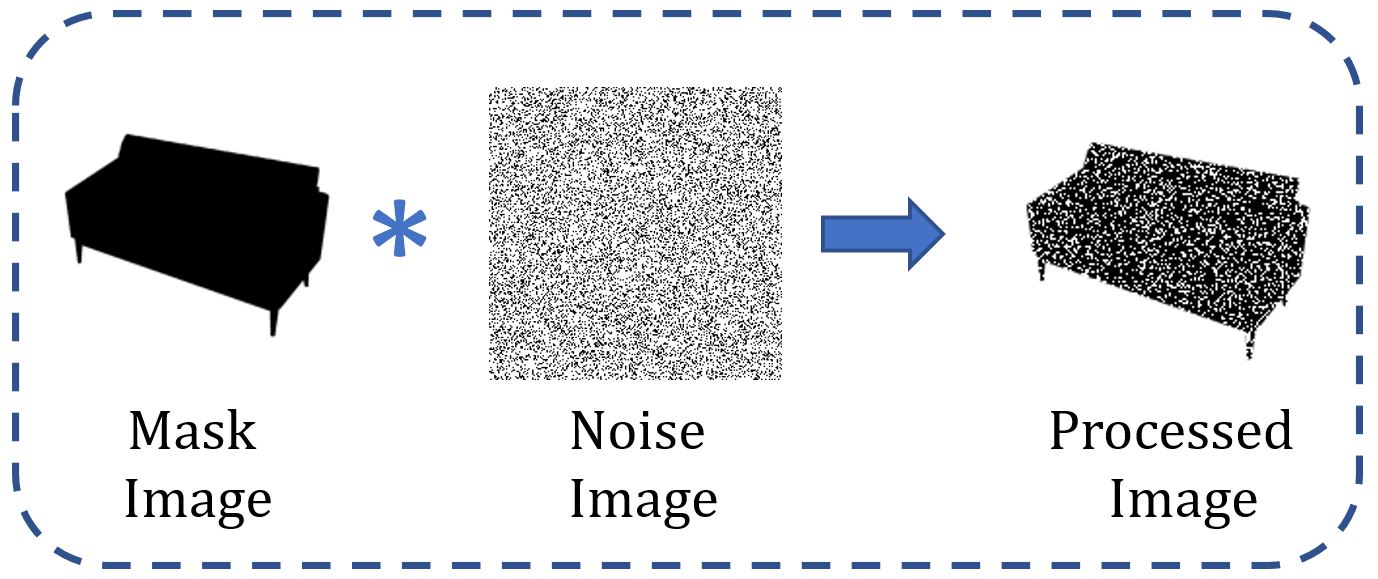} 
\end{center}
  \caption{For a binary mask image, we multiply the binary image with a noise image to the make binary image sparse. The noise image is composed of $50\%$ white pixel and $50\%$ of black pixel sampled randomly.}
\label{fig:preprocess}

\end{figure*}

\textbf{Few-shot Classification:} For few-shot point cloud classification, PointCLIP proposes an inter-view adapter. The inter-view adapter extracts the global visual representation by combining the multi-view features produced by the visual encoder of PointCLIP. The global representation is then added back to the adapted features $F_i^I$. Thus, the adapter can be formulated as follows:
\begin{eqnarray}    
&&G = f_2(ReLU(f_1(concat({F_i^I}_{i=1}^{M}))) \\
&&F^{g} = ReLU(G W^T)\\
&&\widehat{y} = softmax(\sum_i^{M} \alpha_i ((F_i^I + F^{g}) \{P^T\}^T))
\end{eqnarray}
\noindent where $P$ denotes textual information, $\alpha_i$ is a hyper-parameter that denotes importance of view i, $W$ denotes learnable weights, and $f_1$, $f_2$ are MLP layers.

\subsection{Point Cloud to Image Translation Network Pre-training} \label{pre}
Instead of directly applying CLIP \cite{radford2021learning} visual encoder to depth maps, we propose to learn a Point Cloud to Image Translation Network (PCITNet) for aligning point cloud depth features with CLIP visual features. In other words, we expect the extracted features of a rendered point cloud depth map to be consistent with the CLIP visual features of the corresponding image. Then CLIP \cite{radford2021learning} textual prompts can be directly adopted to match the depth features. Let $S =\{B_i,R_i\}_{i=1}^{L}$ denotes a pre-training dataset with $L$ instances. Here $B_i$ is a binary mask image and $R_i$ denotes its corresponding rendered RGB image. We would like to learn a network $F_{\theta}(\cdot)$ that maps from a binary mask image to a rendered RGB image as follows:
\begin{equation}
\widehat{R} = F_{\theta}(B)
\end{equation}

\begin{figure*}[t]
\begin{center}
\includegraphics[width=\linewidth]{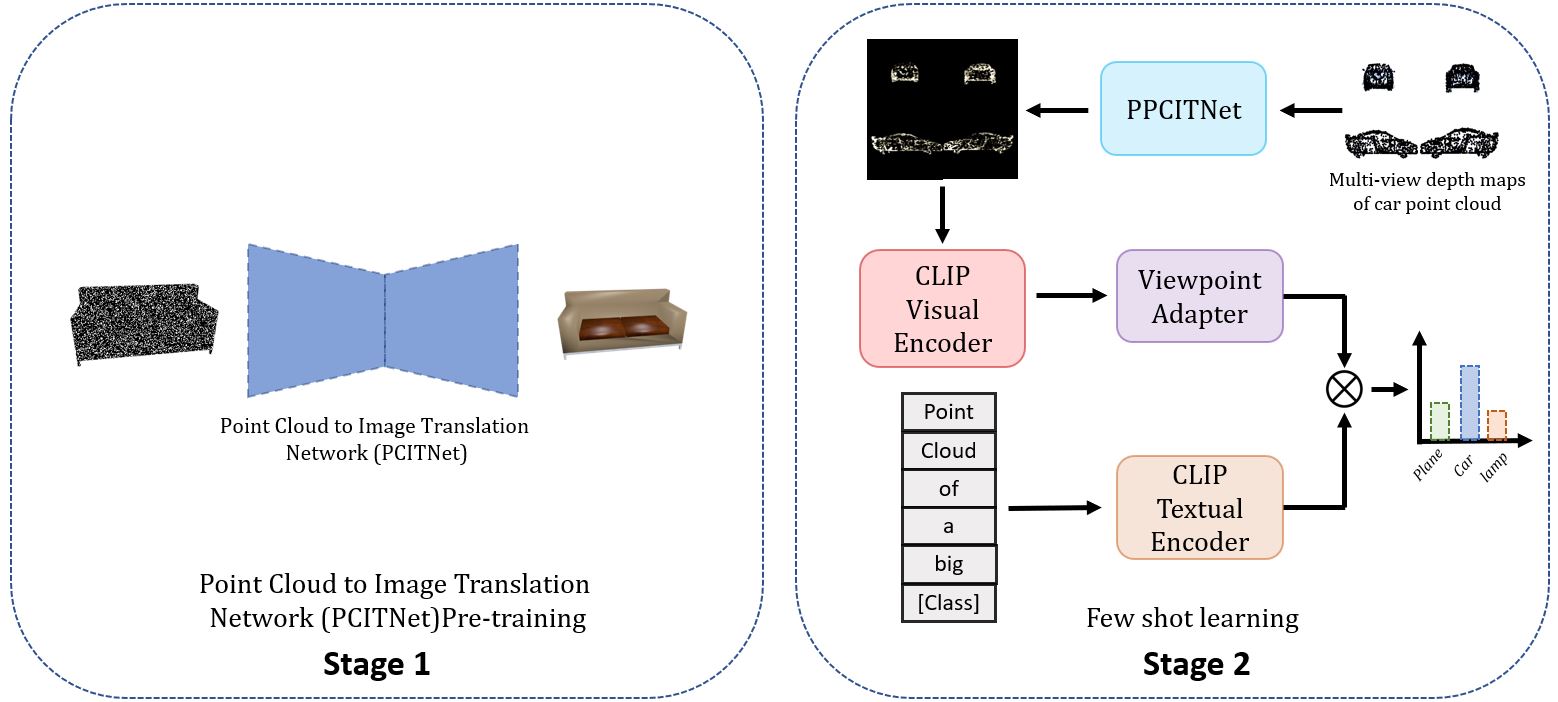} 
\end{center}

  \caption
  {
  The training of our approach is composed of two stages. In the first stage, we pre-train our PCITNet using the processed binary mask and RGB pairs.  In the second stage, we perform a few shot learning on a viewpoint adapter utilizing PPCITNet and pre-trained CLIP.
  }
  
\label{fig:arch}

\end{figure*}

Our goal is to learn the PCITNet $F_{\theta}$ that represents generalized image color distribution along with additional salient visual cues. As discussed earlier,  binary mask images and point cloud depth maps are similar geometrically because of their discrete and compact representations. But visually, they are slightly different. To further bridge the gap, we pre-process a binary mask image by multiplying the binary image with a noise image to make the binary image sparse as depicted in Fig.~\ref{fig:preprocess}. The noise image is composed of $50\%$ white pixel and $50\%$ of black pixel sampled randomly. To learn the generalized image information along with additional salient visual cues, we optimize the following objective function:
\begin{equation}
\mathcal{L}_c = \frac{1}{L}\sum^{L}_{i=1}(R_i-\widehat{R}_i)^2
\end{equation}
Here, $L$ is the total number of mask-RGB pairs in the dataset. The generalized image information along with additional salient visual cues information helps to encode a richer and more diverse set of visual features that can be used to discriminate between different objects. Without image information, CLIP \cite{radford2021learning} may have difficulty distinguishing between objects with similar shapes. For example, consider the task of classifying chairs based on their shape alone. Chairs have similar shape features to tables such as legs. Based on the shape alone, it is very difficult for CLIP \cite{radford2021learning} to distinguish between them. However, by incorporating image information in the classification process, we can identify additional features that can help differentiate between chairs and tables as the image information provides additional cues for the CLIP \cite{radford2021learning} as described by Bramao et. al. \cite{bramao2011role}.

\subsection{Few-Shot Learning} \label{few}

\textbf{Settings.} Let $\rho \in R^{P \times 3}$ denote the point cloud, where $P$ denotes the number of points of the point cloud sample from the $N \times K$ few shot data. Here, $N$ is the total number of classes and each class has $K$ instances of point cloud. Given the PPCITNet and pre-trained CLIP \cite{radford2021learning} network, the goal is to train the viewpoint adapter so that it can boost the performance of the CLIP-based point cloud classification network.

\noindent\textbf{Feature Extraction.} For each $\rho \in R^{P \times 3}$, we need to project 3D coordinates to 2D coordinates. Following \cite{huang2022clip2point}, we get the point cloud depth maps $ f^d=\{f_1,f_2, \dots, f_M\}$. These depth maps are first passed through the PPCITNet, then the output feature is passed through CLIP's visual encoder. The goal of our PPCITNet is to provide generalized image information along with additional salient visual cues to the CLIP model so that it can achieve promising performance on point cloud classification and understanding. 
\begin{eqnarray}
&&f^c = F_{\theta}(f^d),  i= 1,2 \dots M\\
&&f^v = F_{V}(f^c),  i= 1,2 \dots M
\end{eqnarray}
where $i$ indicates the number of depth maps of a 3D point cloud captured from different perspectives, $f_i^v$ denotes output for $f_i^d$ depth map, $f_i^c$ denotes generalized colored images, $F_{\theta}$ and $F_{V}$ denote the PPCITNet and CLIP's visual encoder \cite{huang2022clip2point} respectively.

\noindent \textbf{Viewpoint adapter.} We propose a novel viewpoint adapter that combines the view feature processed by each viewpoint as well as the global intertwined knowledge that exists across the multi-view features. Given the extracted feature $ f^v=\{f_1,f_2, \dots, f_M\}$, the view-specific view information is calculated using $M$ MLP layers. Thus,

\begin{equation}
f_i^{l} = \phi (f_i^v W_{li})
\end{equation}
where $W_{li}$ is the weight of an MLP layer and $\phi$ denotes the activation function. $f_i^{l}$ captures the view-specific fine-grained visual features and generalized image information along with additional salient visual cues that are relevant to a particular point cloud object. For example, to classify the point cloud of an airplane, the viewpoint that contains the wing information is more crucial than any other part. $f_i^{l}$ encodes fine-grained wing information for the point cloud of the airplane. To get the global information of the $M$ views, we perform the following operation:

\begin{equation}
f^{g} = \phi (concat({f_i^v}_{i=1}^{M}) W_{g1}^T) W_{g2}^T
\end{equation}
where $f^{g} \in R^{1 \times C}$ and $W_{g1}$ $W_{g2}$ denote the two-layer weights in the viewpoint adapter. Here, the global knowledge captures the overall structure and organization of point clouds and provides a more holistic understanding of the point cloud objects. Finally, the classification is performed as follows:

\begin{equation}
logits = softmax(\sum_i^{N} \alpha_i ((f_i^{l} + f^{g}) \{P^T\}^T))
\end{equation}
Where $\alpha_i$ denotes the learnable weight, and $P$ denotes textual information. Note that, Only the viewpoint adapter is trained in few-shot learning. The features learned by the viewpoint adapter provide complementary information about the overall structure and view-specific fine-grained features of point cloud objects combining both view and global information.

\section{Experiments} \label{experiment}
\textbf{Pre-training Datasets.} To pre-train our PCITNet network, we use the DISN 2D dataset released by Wang et.al. \cite{xu2019disn}. This dataset is based on the ShapeNet Core dataset \cite{chang2015shapenet}, which is a 3D dataset consisting of 13 object categories. While early work \cite{choy20163d} of rendering this dataset utilizes 24 views with limited variation in terms of camera orientation for each model, DISN provides two types of settings: ``easy'' and ``hard''. The easy setting consists of 36 renderings with smaller variations, The hard setting is composed of 36 renderings with larger variations. To train our  PCITNet network, we sample 100k data from the easy setting randomly. From the RGBA image, we sample the mask image.

\begin{figure*}[t]
\begin{center}
\includegraphics[width=0.65\linewidth]{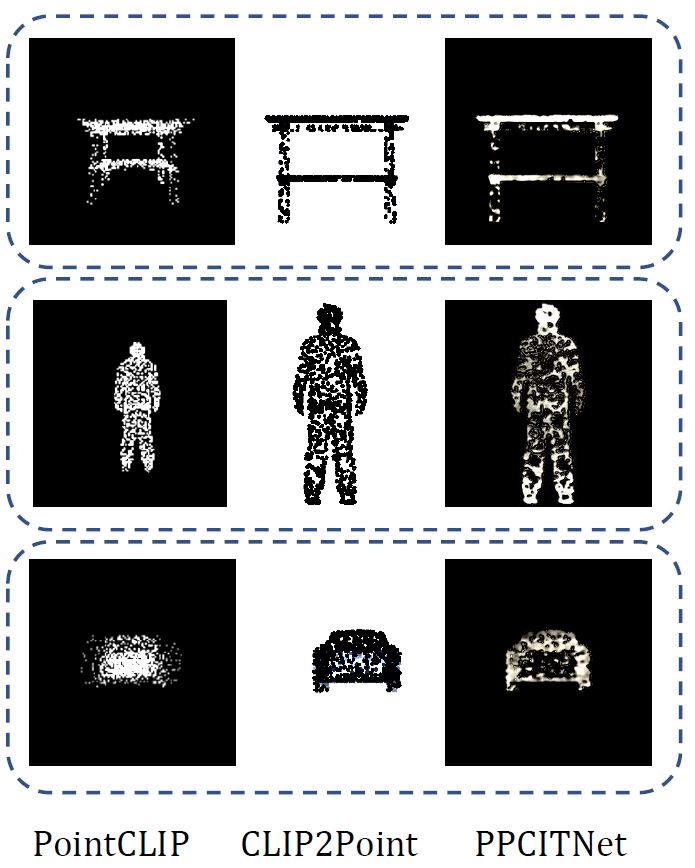} 
\end{center}

  \caption
{
Input Visualization. Our PPCITNet produces generalized colored images along with additional salient visual cues. The salient visual cues refer to additional color concentration to prominent or distinctive parts of the image.
 }
  
\label{fig:visual}
\end{figure*}

\noindent\textbf{Downstream Datasets.} Following PointCLIP \cite{zhang2022pointclip}, we evaluate our proposed model on three widely used benchmark datasets: ModelNet10 \cite{wu20153d}, ModelNet40 \cite{wu20153d} and ScanObjectNN \cite{uy2019revisiting}. ModelNet10 and ModelNet40 have a training point cloud set of  3991 and 9,843 and a test point cloud set of 908 and 2,468 respectively. ScanObjectNN is a real-world point cloud dataset that includes 2,321 samples for training and 581 samples for testing the point cloud from 15 categories. Compare to the ModelNet, ScanObjectNN is more challenging because the CAD models are attached with backgrounds and partially presented. For all three datasets, we uniformly sample 1,024 points of each object  as the PPCITNet's input.

\noindent\textbf{Implementation Details.} We use Unet architecture from \cite{ronneberger2015u} as our  PCITNet network. To pre-train the  PCITNet network, we resize the image to 224 x 224 and train our model in a 12 GB Nvidia Titan X GPU using PyTorch. In pre-training, we use the Adam optimizer \cite{kingma2014adam} with decay of $1 x 10^{-4}$ and the initial learning rate of $1 x 10^{-3}$. Our pre-training task takes 100 epochs with a batch size of 16. For few-shot learning, we utilize AdamW optimizer \cite{loshchilov2017decoupled} with decay of $1 x 10^{-4}$ and the initial learning rate of $1 x 10^{-3}$. The training batch size is 32 and it takes 100 epochs to train the network. Similar to \cite{zhang2022pointclip,huang2022clip2point}, we use the 6 orthogonal views: left, right, top, bottom, front, and back for few-shot learning.

\subsection{Results}
\begin{figure*}[hbt]
\begin{center}
\includegraphics[width=1\linewidth]{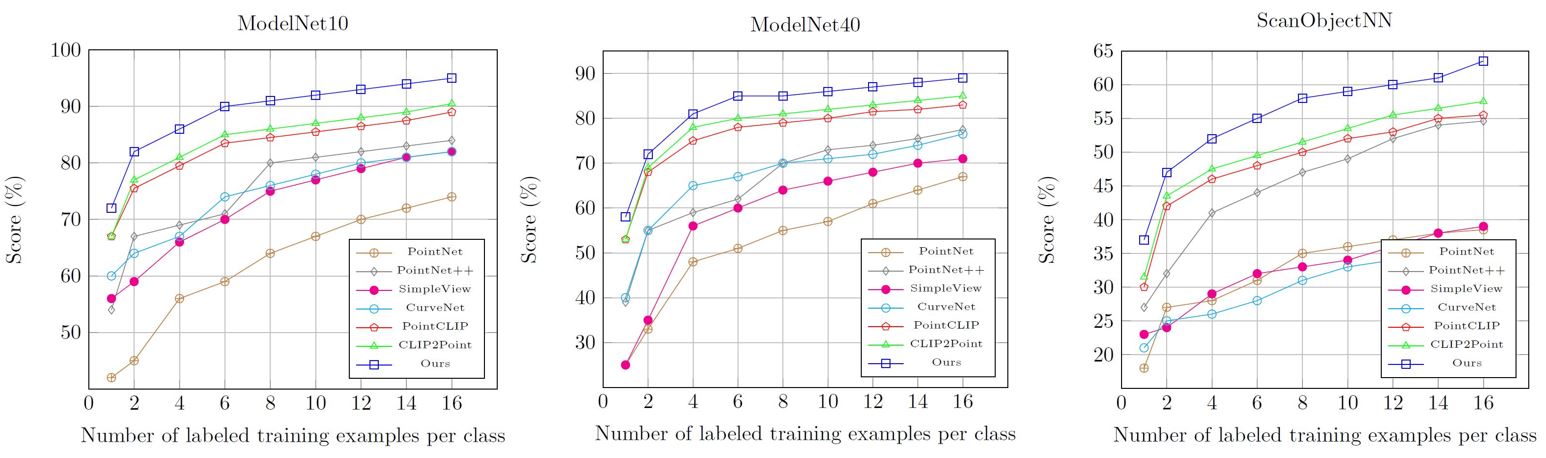} 
\end{center}
  \caption
  {
  Few-shot performance comparison under 1, 2, 4, 8,10,12,14, and 16-shot settings.
  }
  
\label{fig:graph}
\end{figure*}

CLIP-based models \cite{zhang2022pointclip} \cite{huang2022clip2point} are generally evaluated by comparing with state-of-the-art methods on few-shot learning and prompt engineering. In table \ref{tab:tab0}, we present the zero-shot and few-shot performance of PPCITNet on ModelNet40 using the prompt “point cloud of a big [CLASS]”. In table \ref{tab:tab1}, we present the few shot performance of PPCITNet and compare it with state-of-the-art 3D networks like PointNet \cite{qi2017pointnet}, PointNet++ \cite{qi2017pointnet++}, CurveNet \cite{xiang2021walk}, SimpleView \cite{chen2020simple} as well as CLIP based models PointCLIP \cite{zhang2022pointclip}, CLIP2Point \cite{huang2022clip2point} on 16 shot setup. As we can see from the table, PPCITNet with a viewpoint adapter 
outperforms PointCLIP and CLIP2Point by a margin of 3-5 \% for 16 shot setup for prompt ``point cloud of a big [CLASS]" on all three datasets. To further evaluate the transfer ability of PPCITNet, we show the performance for 1,2,4,8,10,12,16 shots in Fig.~\ref{fig:graph}.

\begin{table}[hbt]
    \centering
    \begin{tabular}{c@{\hspace{4em}}c}
        \hline
        \textbf{Setup} & \textbf{Accuracy} \\ \hline
        Zeroshot & 22.74 \\
        Few-shot & \textbf{88.93} \\ \hline
    \end{tabular}
    \caption{Zeroshot and Few-shot results of PPCITNet on ModelNet40 using the prompt “point cloud of a big [CLASS]”.}
    \label{tab:tab0}
\end{table}

We can see from the graph, our PPCITNet surpasses all by a reasonable good margin. This is due to the additional visual cues provided by PPCITNet and the view and global information encoding of the viewpoint adapter. The large performance gain on ScanObjectNN indicates the robustness of PPCITNet under noisy real-world scenes.

\begin{table}[hbt]
\begin{center}
\small 
\setlength{\tabcolsep}{1pt} 
\begin{tabular}{c |c| c| c} 
 \hline
Model & ModelNet10 & ModelNet40 & ScanObjectNN \\ [0.5ex] 
 \hline
CurveNet & 82.45 & 76.55 & 34.76 \\ 
 \hline
SimpleView & 84.15 & 71.17 & 37.44 \\ 
 \hline
PointNet & 73.98 & 67.34 & 36.18 \\ 
 \hline
PointNet++ & 84.62 & 77.13 & 51.62 \\ 
 \hline
PointCLIP & 89.33 & 83.80 & 54.37 \\ 
 \hline
CLIP2Point & 90.21 & 85.10 & 57.49 \\ 
\hline
PPCITNet & \textbf{94.30} & \textbf{88.93} & \textbf{63.22} \\ 
\hline
\end{tabular}
\caption{Performance (\%) of PPCITNet with other methods in 16-shot setup using prompt “point cloud of a big [CLASS]”.}
\label{tab:tab1}
\end{center}
\end{table}

The visualization
in Fig.~\ref{fig:visual} further establishes our claims. While PointCLIP and CLIP2Point provide uniformly sampled point features to the CLIP's visual encoder, our PPCITNet produces generalized colored images along with additional salient visual cues. Here, the salient visual cues refer to additional color concentration to prominent or distinctive parts like an additional color concentration in the head and legs of the human in Fig.~\ref{fig:visual}. In table~\ref{tab:prompt}, we compare our PPCITNet with PointCLIP for different prompt designs on ModelNet40, where [CLASS] represents the class token and `[Learnable Tokens]' refers to the prompts with a fixed length that are capable of being learned during training. The large performance gain indicates the generability of our PPCITNet over PointCLIP for various prompt designs.

\begin{table}[htbp]
\begin{center}
\small 
\setlength{\tabcolsep}{1pt} 
\begin{tabular}{c |c| c} 
 \hline
Prompts & PointCLIP &  PPCITNet \\ [0.5ex] 
 \hline
“a photo of a [CLASS].” & 81.78 & \textbf{86.63} \\ 
 \hline
“a point cloud photo of a [CLASS].” & 82.02 & \textbf{87.33}  \\ 
 \hline
“point cloud of a [CLASS].” & 82.10  &  \textbf{87.04} \\ 
 \hline
“point cloud of a big [CLASS].” & 83.80  & \textbf{88.93} \\ 
 \hline
“point cloud depth map of a [CLASS].” & 81.58  & \textbf{85.15} \\ 
 \hline
“[Learnable Tokens] + [CLASS]” & 69.23 & \textbf{76.27} \\ 
 \hline

\end{tabular}
\caption{Performance (\%) of  PPCITNet with PointCLIP for different prompt designs on ModelNet40. }
\label{tab:prompt}
\end{center}
\end{table}

\subsection{Ablation Studies}
In this section, we evaluate the effect of our PPCITNet and the effect of view information on the viewpoint adapter. To observe the effect of PPCITNet, we conduct an experiment with PPCITNet and without PPCITNet on ModelNet40 as shown in table~\ref{tab:pre-training}. 

\begin{table}[hbt]
    \centering
    \begin{tabular}{cc}
        \hline
        Model & Accuracy \\ \hline
        Without PPCITNet & 84.27 \\
        With PPCITNet & \textbf{88.93} \\ \hline
    \end{tabular}
    \caption{Effect of PPCITNet on ModelNet40 using prompt “point cloud of a big [CLASS]”.}
    \label{tab:pre-training}
\end{table}

From the table, it is evident that incorporating  PPCITNet on the few-shot pipeline improves accuracy by 4.6 \%. To analyze the view feature, we conduct experiments with the only view feature, with only global information, and with both view information and global information on PPCITNet on ModelNet40. Although the performance drops significantly while utilizing only the view information, a combination of view and global information yields the best performance, specifically an improvement of 1.3\% over global information as described in table ~\ref{tab:view}. 

\begin{table}[hbt]
    \centering
    \begin{tabular}{ccc}
        \hline
        View info. & Global info. & Accuracy \\ \hline
        \checkmark & - &  82.34 \\
        - & \checkmark  & 87.60 \\
        \checkmark & \checkmark & \textbf{88.93} \\ \hline
    \end{tabular}
    \caption{Effect of view information for PPCITNet on ModelNet40 using prompt “point cloud of a big [CLASS]”.}
    \label{tab:view}
\end{table}

\section{Conclusion} \label{conclusion}

In conclusion, we present a novel pretrained point cloud to image translation network that transfers image information to the point cloud depth maps. In addition, we present a novel viewpoint adapter that combines the view feature processed by each viewpoint as well as the global intertwined knowledge existing across the multi-view features. The experiment results validate the superior performance of our approach  compared to the other state-of-the-art models on the few-shot point cloud classification.

%
%
%
\bibliographystyle{splncs04}
\bibliography{egbib}

\end{document}